\documentclass[10pt,twocolumn,letterpaper]{article}

\usepackage{wacv}              

\usepackage{graphicx}
\usepackage{amsmath}
\usepackage{amssymb}
\usepackage{booktabs}
\usepackage{multirow}

\usepackage[pagebackref,breaklinks,colorlinks]{hyperref}

\usepackage[capitalize]{cleveref}
\crefname{section}{Sec.}{Secs.}
\Crefname{section}{Section}{Sections}
\Crefname{table}{Table}{Tables}
\crefname{table}{Tab.}{Tabs.}

\def\wacvPaperID{2} 
\def\confName{WACV}
\def\confYear{2025}

\begin{document}

\title{Gate-Shift-Pose: Enhancing Action Recognition in Sports with Skeleton Information}

\author{
Edoardo Bianchi and Oswald Lanz\\
Free University of Bozen-Bolzano\\
Piazza Università 1, Bolzano, Italy\\
{\tt\small \{edbianchi, oswald.lanz\}@unibz.it}
}
\maketitle

\begin{abstract}
This paper introduces Gate-Shift-Pose, an enhanced version of Gate-Shift-Fuse networks, designed for athlete fall classification in figure skating by integrating skeleton pose data alongside RGB frames. We evaluate two fusion strategies: early-fusion, which combines RGB frames with Gaussian heatmaps of pose keypoints at the input stage, and late-fusion, which employs a multi-stream architecture with attention mechanisms to combine RGB and pose features. Experiments on the FR-FS dataset demonstrate that Gate-Shift-Pose significantly outperforms the RGB-only baseline, improving accuracy by up to 40\% with ResNet18 and 20\% with ResNet50. Early-fusion achieves the highest accuracy (98.08\%) with ResNet50, leveraging the model’s capacity for effective multimodal integration, while late-fusion is better suited for lighter backbones like ResNet18. These results highlight the potential of multimodal architectures for sports action recognition and the critical role of skeleton pose information in capturing complex motion patterns. Project page: https://edowhite.github.io/Gate-Shift-Pose
\end{abstract}

\section{Introduction}
\label{sec:intro}
Human action recognition in sports presents unique challenges, including rapid motion, frequent occlusions, and complex postures \cite{sportActionRecSurvey}. While earlier approaches primarily relied on single-modal inputs, such as RGB frames, recent advancements have shifted toward multi-modal inputs to better capture the intricate spatial and temporal dynamics of athletic movements \cite{ActionRecSurvey}. This evolution is particularly crucial for complex sports like figure skating, where fast movements, intricate maneuvers, and low-to-ground rotations create significant challenges for accurately recognizing actions.

Building on these advancements, we introduce Gate-Shift-Pose (GSP), an extension of Gate-Shift-Fuse (GSF) networks \cite{GSF}, designed to address the specific challenges of action recognition in dynamic contexts. GSF was chosen as the foundational architecture for its proven effectiveness and versatility across various domains, including sports \cite{GSF} and industrial applications \cite{EgoActionRec}. By incorporating skeleton-based pose information through multiple fusion strategies, GSP integrates complementary visual and structural features, offering a more robust and precise representation of athletic actions.

We evaluate our approach using the FR-FS \cite{FRFS} ice skating dataset, specifically designed for fall classification. Our experimental results demonstrate that incorporating pose information improves classification accuracy, highlighting the effectiveness of multimodal architectures in sports action recognition.

The main contributions of this paper are summarized below:
\begin{itemize}
    \item We introduce Gate-Shift-Pose (GSP), a novel extension of the Gate-Shift-Fuse (GSF) network that incorporates skeleton-based pose information for enhanced fall classification.
    
    \item We propose an early-fusion approach that integrates pose data as a Gaussian heatmap channel alongside RGB frames, allowing the model to leverage spatial and structural information from the input stage.
    
    \item We propose a late-fusion approach that employs a two-stream network with multi-head attention, enabling selective focus on relevant pose and RGB features for improved classification.
\end{itemize}

This work is organized as follows: \cref{sec:background} covers related works, \cref{sec:methods} details the GSP architecture, \cref{sec:results} presents experiments and results, \cref{sec:limitations} discusses limitations, and \cref{sec:discussion} concludes with insights and future work.

\section{Background and Related Work}
\label{sec:background}

Action recognition has gained significant interest in computer vision, particularly in individual and teams sports, where complex dynamics and high-speed movements create unique challenges \cite{ActionRecSurvey, TeamSportActionRecSurvey}. This section reviews traditional and deep learning-based methods for action recognition, focusing on advancements that address these complexities.

\subsection{Traditional Models for Action Recognition}
Traditional action recognition models rely on hand-crafted features, typically following a two-stage pipeline: feature extraction from video frames and subsequent classification. Key techniques in feature extraction include methods like GIST \cite{gist} and Histogram of Oriented Gradients (HOG) \cite{hog}, which capture spatial structures in frames. Kuehne et al. \cite{Kuehne2011} demonstrated that GIST is especially effective on datasets such as UCF Sports \cite{ucfSport} due to its ability to capture contextual scene information. Extensions of HOG, like HOG3D \cite{HOG3D}, enhance action recognition by incorporating spatio-temporal dynamics, delivering promising results on sports-focused datasets. Despite these successes, traditional models often lack end-to-end learning capabilities, as feature extraction and classification are handled separately, which limits their adaptability and efficiency in complex scenarios \cite{sportActionRecSurvey}.

\subsection{Deep Learning Models for Action Recognition}
Deep learning has advanced action recognition considerably by enabling end-to-end training, which improves the model's ability to generalize across various sports actions. Several prominent categories of deep learning models have emerged:

\paragraph{2D CNN Models}
2D CNNs represent early developments in deep learning for action recognition, where individual frames are processed independently before fusing the extracted features over time. Karpathy et al. \cite{Karpathy2014} explored different fusion strategies, including single-frame, early-fusion, late-fusion, and slow-fusion, with slow-fusion demonstrating the best performance on large-scale datasets like Sports 1M \cite{Karpathy2014}.

\paragraph{3D and Pseudo-3D CNN Models}
3D CNNs, such as C3D by Tran et al. \cite{Tran2015} and the Inflated 3D CNN (I3D) proposed by Carreira and Zisserman \cite{Carreira2017}, extend the CNN architecture to both spatial and temporal dimensions, allowing for simultaneous extraction of spatio-temporal features. These models perform well on datasets such as Sports 1M \cite{Karpathy2014}, where complex movements and high-speed actions are common, demonstrating the effectiveness of 3D CNNs for sports-based action recognition.

Pseudo-3D models approximate 3D convolutions by decoupling spatial and temporal processing, achieving greater efficiency. Examples include the Temporal Segment Network (TSN) \cite{TSN}, which uses sparsely sampled frames to capture long-term dependencies, and the Temporal Shift Module (TSM) \cite{TSM}, which shifts feature maps temporally within a 2D CNN, reducing the need for full 3D convolutions. Gate-Shift-Fuse (GSF) \cite{GSF} builds on these by introducing gated temporal shifts and adaptive fusion mechanisms, effectively capturing motion features. Together, these pseudo-3D approaches provide a computationally efficient and effective solution for large-scale sports action recognition.

\paragraph{Two-Stream Models}
Two-stream networks are designed to separate appearance and motion by processing RGB frames and optical flow independently. Works by Simonyan et al. \cite{Simonyan2014} and C. Feichtenhofer et al. \cite{Feichterhofer2016} established the two-stream architecture, which has shown high efficacy in sports contexts by combining spatial and motion features. This separation allows the model to capture intricate motion patterns while preserving appearance details.

\paragraph{Skeleton-Based Models}
Skeleton-based methods employ Graph Convolutional Networks (GCNs) \cite{GCN} to model joint movements, focusing on skeletal properties rather than appearance. The Spatial-Temporal Graph Convolutional Network (ST-GCN) introduced by Yan et al. \cite{Yan2018} has significantly contributed to advancements in sports action recognition, where capturing precise joint dynamics is crucial. Although these models do not incorporate appearance information, they offer a computationally efficient alternative for sports applications by leveraging skeletal structure alone.

A key limitation of skeleton graph-based models is their dependence on precise joint detection, as inaccuracies can result in noisy graphs that degrade performance on existing datasets. PoseC3D \cite{PoseC3D} tackles this problem by using heatmaps of joints and limbs instead of skeleton graphs, which improves robustness. By treating these heatmaps as frames, PoseC3D can effectively leverage traditional 3D convolutional networks.

RNN-based approaches provide an alternative for modeling sequential data in skeleton-based action recognition. Recurrent Neural Networks (RNNs) \cite{rnn}, particularly Long Short-Term Memory (LSTM) networks \cite{lstm}, are effective at capturing temporal dependencies across joint positions, enabling the modeling of motion patterns over time. Examples include the work by Du et al. \cite{HRNN} and Lee et al. \cite{Lee2017}.

\subsection{Current Challenges in Individual Sports Action Recognition}
Deep learning has advanced action recognition, but challenges remain, especially in sports with dynamic interactions and complex poses. Key issues include occlusions, high-speed movements, and diverse environments, highlighting the need for further innovations to enhance robustness and accuracy \cite{sportActionRecSurvey, ActionRecSurvey}.

\begin{figure*}[ht]
    \centering
    \includegraphics[width=\textwidth]{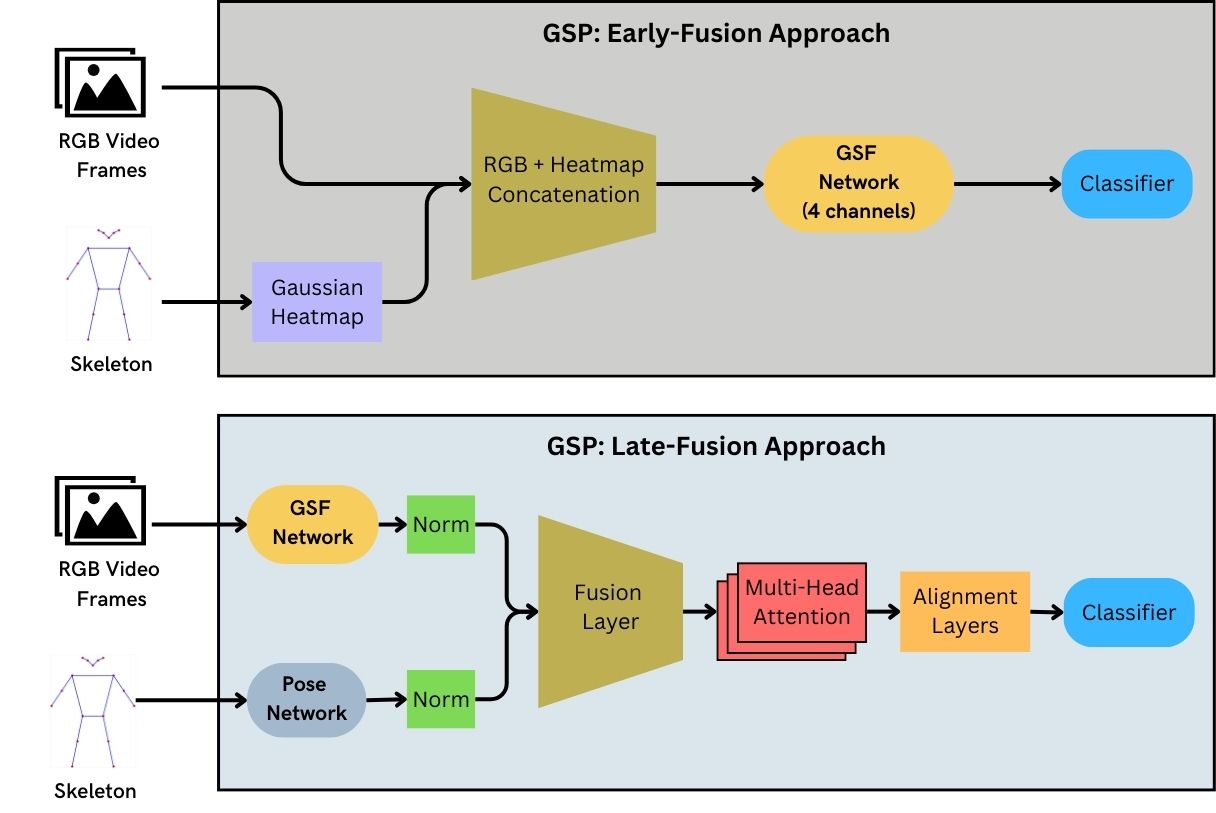}
    \caption{Overview of the GSP (Gate-Shift-Pose) network architecture with two fusion strategies for integrating RGB and skeletal information. \textbf{Top:} In the early-fusion approach, pose data is preprocessed as a Gaussian heatmap and concatenated with RGB frames, forming a four-channel input for the GSF network. \textbf{Bottom:} In the late-fusion approach, RGB frames and skeletal data are processed in separate streams using a GSF network and a Pose network, respectively. Normalized features from each stream are then combined in a fusion layer, followed by multi-head attention and alignment layers to integrate relevant spatio-temporal features before classification.}
    \label{fig:fusion_approaches}
\end{figure*}

\section{Proposed Methodology}
\label{sec:methods}
This section details Gate-Shift-Pose (GSP), our proposed multimodal architecture, which integrates pose information into Gate-Shift-Fuse (GSF) networks for sport action classification. We outline the original GSF module in \cref{sec:gsf}, present the skeleton pose estimation framework in \cref{sec:poses}, and discuss the two fusion strategies used to combine pose and RGB information in Sections \ref{sec:ef} and \ref{sec:lf}, respectively.

\subsection{Gate-Shift-Fuse Networks}
\label{sec:gsf}
The Gate-Shift-Fuse (GSF) module \cite{GSF} enhances 2D CNNs for efficient spatio-temporal learning in video tasks by integrating spatial gating and weighted channel fusion with temporal shifts. Inspired by the Temporal Segment Network (TSN) \cite{TSN} and Temporal Shift Module (TSM) \cite{TSM}, GSF adapts each layer to process temporal dependencies with minimal overhead. It consists of a spatial gating block that selectively routes features across frames, and a fusion block that merges shifted temporal features with residual spatial information.

The GSF module first applies a 2D convolution to capture spatial features, then splits these features by channel and modulates them through a 3D convolution-based gating layer. A temporal shift operation adds temporal dependencies, and the features are recombined through a channel-wise fusion layer, enabling adaptive spatio-temporal representation.

\begin{figure*}[ht]
    \centering
    \includegraphics[width=\textwidth]{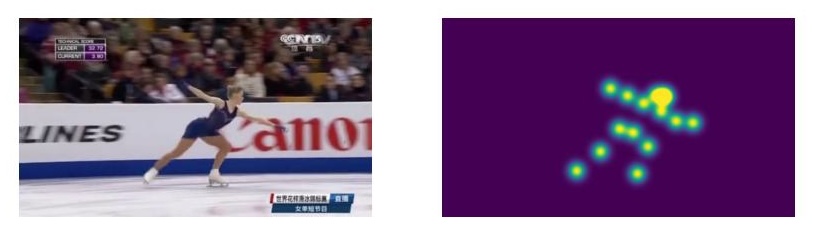}
    \caption{Example from the FR-FS dataset. \textbf{Left:} an RGB frame of an ice skater performing a maneuver. \textbf{Right:} the corresponding Gaussian heatmap highlighting keypoints for skeleton-based feature extraction.}
    \label{fig:frfs_example}
\end{figure*}

\subsection{Pose Estimation Framework}
\label{sec:poses}
The pose estimation framework is a critical component in our approach, as it provides structural information that enhance the network's ability to identify fall events by detecting and localizing specific body keypoints. We employ the YOLO11-x-pose model \cite{yolo11}, pretrained on the COCO Keypoints dataset \cite{coco}, which is capable of detecting 17 anatomical keypoints corresponding to human features. These keypoints include the nose, eyes, ears, shoulders, elbows, wrists, hips, knees, and ankles.

The YOLO11-x-pose model achieves high accuracy in pose estimation tasks, with a mean Average Precision (mAP) of 69.5\% at an Intersection over Union (IoU) threshold of 0.5:0.95, and 91.1\% at an IoU threshold of 0.5. These results highlight the model's robustness and reliability in accurately detecting body postures. The model comprises 58.8 million parameters.

To incorporate pose information into our architecture, we precompute and store the 17 keypoint coordinates for each frame in the dataset. This pose data is saved on disk and loaded during both training and inference. Depending on the fusion strategy—early-fusion or late-fusion—the poses are either transformed into Gaussian heatmaps and concatenated with the RGB channels or processed through a separate network stream dedicated to pose information.

\subsection{GSP: Early-Fusion Skeleton Integration}
\label{sec:ef}
In the early-fusion approach, presented in Fig. \ref{fig:fusion_approaches} (Top), we incorporate pose information by augmenting each RGB frame with an additional channel containing the Gaussian heatmap of pose keypoints. This results in a four-channel input (RGB + pose) per frame, which is processed by a GSF network. This approach allows the network to learn correlations between pose and appearance features at an early stage, potentially capturing valuable low-level interactions. An example of a video frame from the FR-FS dataset with the respective Gaussian heatmap is shown in Fig. \ref{fig:frfs_example}.

Early-fusion is computationally efficient, as both RGB and pose information are processed jointly by the same feature extractor. However, by fusing the modalities from the initial layers, this approach may be limited in capturing higher-level interactions between the pose and RGB features.

\begin{figure}[ht]
    \centering
    \includegraphics[width=\linewidth]{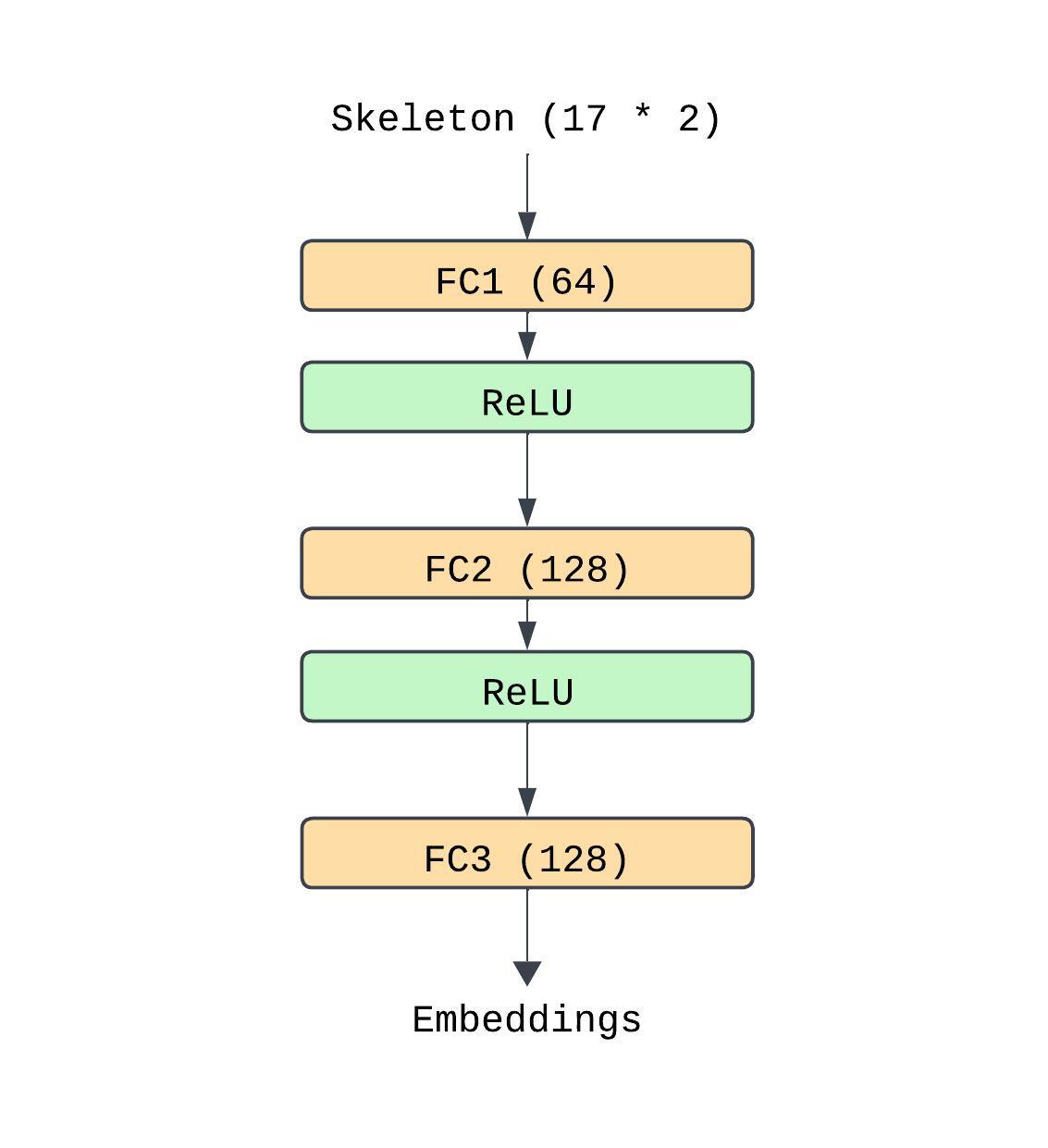}
    \caption{Architecture of the pose model. The model processes a skeleton input consisting of 17 keypoints, each represented by \textit{x} and \textit{y} coordinates (17 x 2). The input is passed through three fully connected (FC) layers: FC1 (64 neurons), FC2 (128 neurons), and FC3 (128 neurons). ReLU activation functions are applied after the first two layers. The output is a feature embedding vector suitable for downstream tasks.}
    \label{fig:posemodel}
\end{figure}

\subsection{GSP: Late-Fusion Skeleton Integration}
\label{sec:lf}
The late-fusion strategy, illustrated in Fig. \ref{fig:fusion_approaches} (Bottom), employs a two-stream architecture with separate branches for RGB frames and poses. The RGB stream processes raw visual data with a GSF network, while the pose stream leverages a dedicated MLP-based model to embed joint coordinates into feature vectors. As presented in \cref{fig:posemodel}, the pose model consists of three fully connected layers: the first layer maps the 34-dimensional input (representing 17 joints, each with x and y coordinates) to a 64-dimensional space, followed by an intermediate layer expanding to 128 dimensions, and a final layer outputting a 128-dimensional embedding. ReLU activations are applied after each layer, resulting in a compact representation of skeletal dynamics.

Following independent feature extraction, the RGB and pose features undergo L2 normalization to ensure balanced scaling between modalities. The normalized features are then concatenated to form a combined representation, which is subsequently processed by a multi-head attention layer. This attention mechanism allows the model to dynamically emphasize contextually relevant features across the RGB and pose streams, thereby enhancing its sensitivity to critical aspects of each modality within the fused representation.

\begin{figure}[ht]
    \centering
    \includegraphics[width=\linewidth]{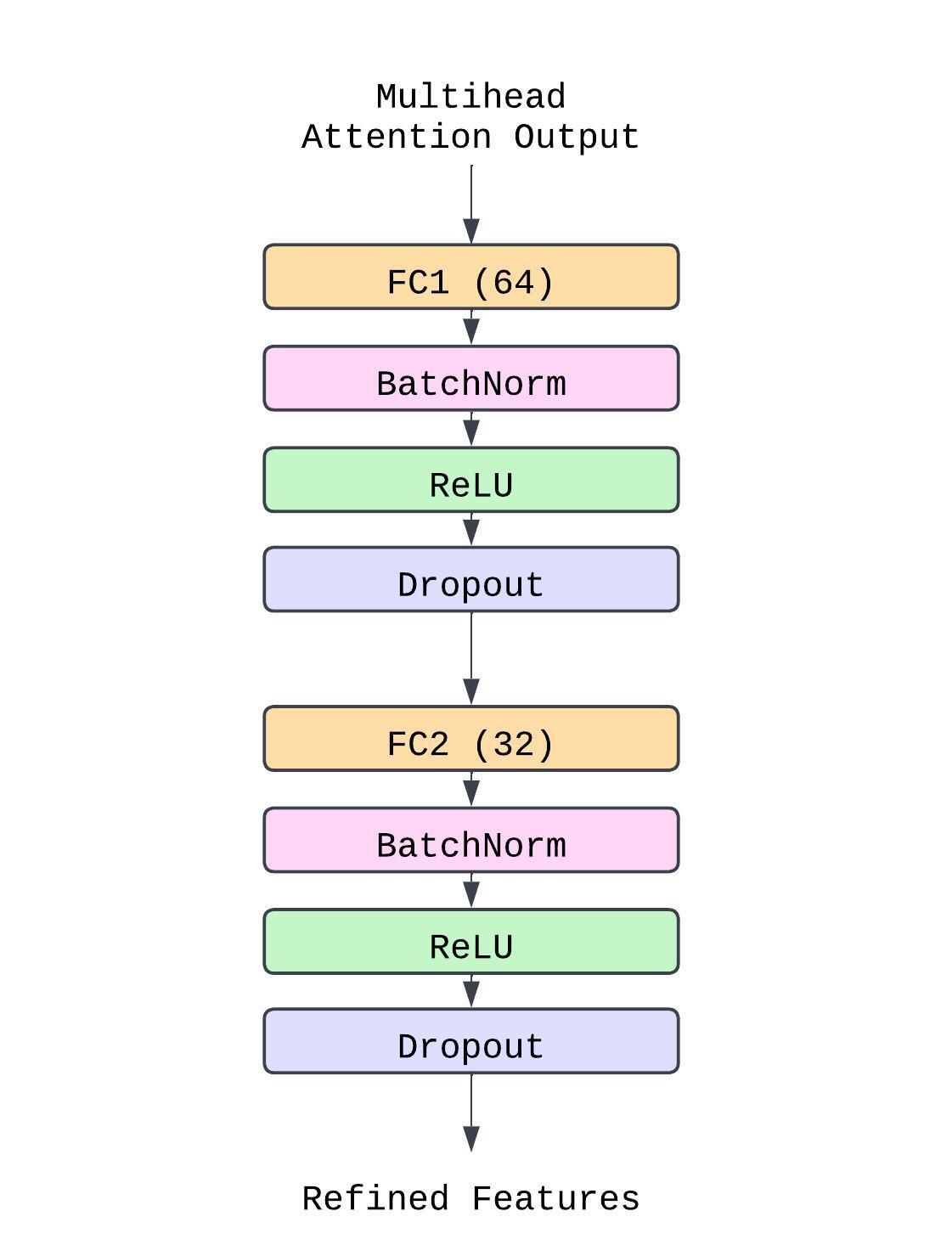}
    \caption{Alignment layers applied after the Multihead Attention module and before the classification layer. These layers are designed to compress and select the most relevant information for the downstream task. The structure includes two fully connected layers (FC1 with 64 neurons and FC2 with 32 neurons), each followed by Batch Normalization, ReLU activation, and Dropout to enhance generalization and robustness.}
    \label{fig:alignlayers}
\end{figure}

The attention-enhanced output is further refined by a feature refinement module, designed to compress and selectively emphasize relevant information. This module, as presented in \cref{fig:alignlayers}, consists of a sequential network of two linear layers: the first layer reduces the feature dimension by half, followed by batch normalization, ReLU activation, and dropout. A second linear layer then reduces the dimensionality by another factor of two, followed by batch normalization, ReLU, and dropout again. This structure provides a form of regularization, compressing the fused features and mitigating noise, leading to a more robust and discriminative feature set. The final representation is passed through a fully connected layer for classification. 

The late-fusion approach enables the model to effectively leverage complementary RGB and pose information, enhancing its ability to distinguish complex movements. Although this architecture introduces additional computational overhead, it facilitates contextually aware interactions between the modalities, which is particularly advantageous for recognizing complex and dynamic actions in sports like figure skating.

\section{Experiments and Results}
\label{sec:results}
We evaluated our proposed multimodal approach on the FR-FS ice skating dataset, specifically designed for binary fall classification. Our evaluation compared three networks: the baseline GSF without pose information, GSP with early-fusion, and GSP with late-fusion. In both the early- and late-fusion setups, skeleton pose information was integrated to enhance fall detection capabilities by leveraging both visual and structural features.

To further analyze model performance and generalizability, we employed two different backbone networks for each GSF/GSP variant: ResNet18 \cite{resnet} and ResNet50 \cite{resnet}. This dual-backbone approach allowed us to assess the impact of model complexity on classification accuracy, particularly in the context of high-speed and intricate actions characteristic of figure skating.

\subsection{Dataset}
The FR-FS dataset \cite{FRFS} comprises 417 video samples, including 276 positive (fall) and 141 negative (non-fall) samples, each containing 103 frames. These videos were collected from the FIV dataset and footage from the PyeongChang 2018 Winter Olympic Games. The dataset captures critical movements during an athlete's take-off, rotation, and landing, providing a comprehensive basis for analyzing the detailed motion dynamics essential for accurate fall classification.

\subsection{Implementation Details}
All model backbones were initialized with ImageNet \cite{imagenet} weights and trained using Stochastic Gradient Descent (SGD) with a momentum of 0.9 and a weight decay of 5e-4. The initial learning rate was set to 0.01 and decayed gradually following a cosine annealing schedule. We experimented with batch sizes of 4 or 8, and segment counts of 16 or 32, training each model variant for 120 epochs. Pose information was incorporated either as a Gaussian heatmap channel in the early fusion strategy or as a dedicated pose stream in the late-fusion strategy.

\subsection{Evaluation Criteria}
We evaluated model performance on the FR-FS dataset using classification accuracy, which quantified the model’s ability to distinguish falls from non-falls. Accuracy was analyzed across different fusion strategies (early- and late-fusion) and backbone architectures (ResNet18 and ResNet50) to assess the effectiveness of each configuration in handling the complex motions typical of figure skating. We also compared with the original (RGB-only) GSF to assess the importance of skeletal information in enhancing classification performance.

\subsection{Results}
Table \ref{tab:results} summarizes the performance of Gate-Shift-Pose on the FR-FS dataset across different fusion strategies, backbones, batch sizes, and segment counts. The highest accuracy (98.08\%) was achieved with early-fusion on ResNet50 with a batch size of 4 and 32 segments, highlighting that integrating RGB and pose data at the input stage is highly effective for models with greater capacity. For ResNet18, the best result (95.19\%) was achieved with late-fusion using a batch size of 4 and 32 segments, demonstrating that maintaining separate RGB and pose streams until later stages is advantageous for lighter backbones.

The integration of pose data in GSP consistently outperformed the original RGB-only GSF baseline, underscoring the value of pose information for classification. Specifically, the inclusion of skeleton-based features improved accuracy from 67.79\% to 95.19\% with ResNet18 (an increase of approximately 40\%) and from 81.73\% to 98.08\% with ResNet50 (an increase of approximately 20\%). These results confirm the effectiveness of pose information in capturing complex motion patterns.

Furthermore, a batch size of 4 resulted in improved performance due to its stabilizing effect on training with relatively small datasets \cite{smallBatch}. Segment count also influenced performance: a higher number of segments (32) consistently led to better results across both ResNet18 and ResNet50, likely due to the richer temporal context enabling more effective modeling of dynamic motion patterns.

In summary, early-fusion is well-suited for larger backbones, enabling effective multimodal integration, whereas late-fusion better supports smaller backbones by reducing stream complexity. Batch size and segment count further modulate performance, indicating the importance of aligning these parameters with model capacity and task-specific requirements to achieve optimal results.

\begin{table*}[ht]
\centering
\caption{Classification accuracy for different GSP fusion strategies, batch sizes, and segment counts, grouped by backbone. Best results for each backbone are in bold. All models were trained with a learning rate of 0.01 and for 120 epochs.}
\label{tab:results}
\begin{tabular}{|c|c|c|c|c|}
\hline
\textbf{Backbone} & \textbf{Method} & \textbf{Batch Size} & \textbf{Segments} & \textbf{Accuracy (\%)} \\
\hline
\hline
\multirow{12}{*}{ResNet18} & \multirow{4}{*}{GSF Baseline} & 4  & 16 & 66.83 \\
                           &                                & 4  & 32 & 67.31 \\
                           &                                & 8  & 16 & 66.34 \\
                           &                                & 8  & 32 & 67.79 \\
                           \cline{2-5}
                           & \multirow{4}{*}{GSP Early-Fusion} & 4  & 16 & 67.31 \\
                           &                                    & 4  & 32 & 66.35 \\
                           &                                    & 8  & 16 & 77.40 \\
                           &                                    & 8  & 32 & 81.25 \\
                           \cline{2-5}
                           & \multirow{4}{*}{GSP Late-Fusion}  & 4  & 16 & 90.39 \\
                           &                                   & 4  & 32 & \textbf{95.19} \\
                           &                                   & 8  & 16 & 87.17 \\
                           &                                   & 8  & 32 & 89.90 \\
\hline
\hline
\multirow{12}{*}{ResNet50} & \multirow{4}{*}{GSF Baseline} & 4  & 16 & 70.68 \\
                           &                                & 4  & 32 & 75.00 \\
                           &                                & 8  & 16 & 73.55 \\
                           &                                & 8  & 32 & 81.73 \\
                           \cline{2-5}
                           & \multirow{4}{*}{GSP Early-Fusion} & 4  & 16 & 93.27 \\
                           &                                    & 4  & 32 & \textbf{98.08} \\
                           &                                    & 8  & 16 & 97.12 \\
                           &                                    & 8  & 32 & 95.19 \\
                           \cline{2-5}
                           & \multirow{4}{*}{GSP Late-Fusion}  & 4  & 16 & 85.10 \\
                           &                                   & 4  & 32 & 82.21 \\
                           &                                   & 8  & 16 & 86.53 \\
                           &                                   & 8  & 32 & 87.02 \\
\hline
\end{tabular}
\end{table*}

\section{Limitations}
\label{sec:limitations}
While our approach shows significant improvements in fall classification accuracy, several limitations should be noted. First, in real-time scenarios, the reliance on pose estimation could introduce latency, as pose calculations may be computationally intensive. This issue can be mitigated by using a lighter pose estimation model, such as YOLO11-n-pose \cite{yolo11} or YOLO11-s-pose \cite{yolo11}, which could reduce latency to acceptable levels for real-time deployment.

High-quality pose data is also essential for optimal performance, making this approach more suited to individual sports where occlusions are minimal, and the athlete remains clearly visible in the frame. In contexts with frequent occlusions or complex multi-person interactions, pose estimation accuracy could decrease, impacting overall model performance.

Finally, the FR-FS dataset contains a limited number of samples, which may affect the model's generalizability, particularly for diverse or extreme motion scenarios that are likely in other sports.

\section{Discussion and Conclusion}
\label{sec:discussion}
In this work, we introduce Gate-Shift-Pose (GSP), an enhanced version of Gate-Shift-Fuse (GSF) networks that incorporates skeletal pose information for improved fall classification in figure skating. Specifically, we explore two distinct fusion strategies: early-fusion, where pose data is integrated as Gaussian heatmaps alongside RGB frames, and late-fusion, which utilizes a two-stream architecture with multi-head attention to effectively combine RGB and pose features.

Our experimental results demonstrate that GSP outperforms the original RGB-only GSF model, highlighting the critical value of pose information for accurately capturing complex movement patterns. Early-fusion achieves the best performance with ResNet50, while late-fusion shows advantages with ResNet18, suggesting that the selection of fusion strategy should align with model capacity to optimize performance.

Furthermore, we observed that parameter choices such as batch size and segment count significantly impact performance. Smaller batch sizes and a higher number of segments improved temporal coverage and stability, which are particularly beneficial for training on this relatively small dataset.

In conclusion, GSP demonstrates the potential of multimodal architectures for complex sports action recognition, emphasizing the importance of aligning fusion strategy with backbone capacity. Future work will focus on optimizing pose estimation for real-time applications and expanding to larger, more diverse datasets to enhance robustness and generalizability across sports.




{\small
\bibliographystyle{ieee_fullname}
\bibliography{egbib}
}

\end{document}